\documentclass[12pt]{article}

\usepackage{amssymb}
\usepackage{amsmath}
\usepackage{multirow}
\usepackage{hyperref}
\usepackage{url}
\usepackage{graphicx}
\usepackage{caption}
\usepackage{subcaption}
\usepackage{booktabs}
\usepackage{array}
\usepackage{xcolor}
\usepackage{makecell}
\usepackage{tabularx}
\usepackage{threeparttable}
\usepackage{adjustbox}
\usepackage{xurl}
\usepackage{tikz}
\usetikzlibrary{positioning,arrows.meta,calc}

\newcolumntype{C}[1]{>{\centering\arraybackslash}p{#1}}

\makeatletter
\def\ps@pprintTitle{%
     \let\@oddhead\@empty
     \let\@evenhead\@empty
     \def\@oddfoot{}%
     \let\@evenfoot\@oddfoot}
\makeatother

\begin{document}

\title{PlagiaCorpus: A Bilingual, Adversarial, and Cross-Lingual Dataset for Forensic Plagiarism Detection}

\author{
    \mdseries Adel Ammar\textsuperscript{1}\textsuperscript{*}, Serry Sibaee\textsuperscript{1},
    Wadii Boulila\textsuperscript{1} \\
    \\
    \textsuperscript{1}Robotics and Internet-of-Things Laboratory (RIOTU),\\ Prince Sultan University, Riyadh 11586, Saudi Arabia \\
    \textsuperscript{*}Corresponding author: \texttt{aammar@psu.edu.sa}
}

\date{}

\maketitle

%% Abstract
\begin{abstract}
PlagiaCorpus is a bilingual Arabic--English dataset for benchmarking plagiarism and text-reuse detection, including reuse produced by large language models (LLMs) and reuse across languages. It contains 198,643 text records: a source corpus of 34,885 passages drawn from English and Arabic Wikipedia and two openly licensed book collections, and 163,758 queries (147,381 for training and 16,377 for testing). The queries cover five reuse classes of increasing difficulty, namely verbatim copying, character-level obfuscation, same-language paraphrasing by two different LLMs, and English--Arabic translation with structural paraphrasing, together with a Safe class of non-reused passages. Every positive query is linked to its source passage, and the train/test split is grouped by source passage so that no passage contributes queries to both splits. The release was curated through language identification, removal of LLM generation artefacts, and exact and cross-lingual de-duplication. Technical validation reports per-class length statistics, automatic checks of the train/test split, and inter-annotator agreement on a manually rated sample. The dataset is distributed in JSON and Parquet formats under a CC BY-SA 4.0 license.
\end{abstract}

\section*{Background \& Summary}
Plagiarism detection is a critical area in natural language processing (NLP), requiring standardized benchmark datasets for the reproducible evaluation of detection methods. The field has moved from string-matching techniques to semantic analysis driven by deep learning and large language models (LLMs). Historically, the corpora of the PAN plagiarism detection competitions~\cite{potthast2009overview,potthast2010overview,potthast2011pan,potthast2013overview} served as the standard benchmarks, progressively introducing more realistic obfuscation strategies and cross-language cases. The PAN plagiarism detection task was revived in 2025 with a dataset of generative plagiarism in scientific articles, created with Llama, DeepSeek-R1 and Mistral models~\cite{greiner2025tap}. The task was continued at PAN 2026 as a retrieval-based challenge, in which a suspicious document generated by an LLM from one or more source documents must be matched to its sources within a candidate collection~\cite{pan2026generative}.

However, the challenge of plagiarism detection is amplified in languages with complex morphology, such as Arabic. Unlike English-centric datasets, Arabic benchmarks must contend with highly derivational word forms, making traditional string-matching ineffective \cite{bensalem2015araplagdet}. Furthermore, recent generative plagiarism datasets such as that of PAN 2025 are English-only and restricted to scientific articles from arXiv, which limits their use for Arabic and cross-lingual scenarios.

PlagiaCorpus addresses these limitations by combining Arabic and English. It contains LLM-generated paraphrases in both languages as well as English-to-Arabic and Arabic-to-English reuse, a combination not covered by existing corpora. It also contains character-level obfuscations (leetspeak substitutions for English and tatweel insertion for Arabic) that target detectors relying on exact character or token overlap, and Safe passages that allow detectors to be evaluated on false positives as well as on detection.

As detailed in Table \ref{tab:dataset_comparison}, PlagiaCorpus bridges the gap between traditional forensic plagiarism corpora and emerging generative benchmarks. Earlier resources such as PAN-PC-11~\cite{potthast2011pan} and AraPlagDet~\cite{bensalem2015araplagdet} incorporate paraphrasing and artificial obfuscation, but predate LLM-generated plagiarism; PAN-PC-11 includes cross-language cases but no Arabic, and AraPlagDet is Arabic-only. PAWS~\cite{zhang2019paws}, although designed around challenging paraphrase and non-paraphrase pairs, is English-only and does not address cross-lingual or LLM-generated plagiarism. More recent benchmarks such as PAN 2026~\cite{pan2026generative} extend plagiarism detection to LLM-generated content, but remain English-only. PlagiaCorpus is the only dataset in this comparison that combines Arabic coverage, cross-lingual reuse and LLM-generated plagiarism.

\begin{table*}[!htbp]
\centering
\caption{Comparison of Forensic Plagiarism and Paraphrase Datasets}
\label{tab:dataset_comparison}
\begin{adjustbox}{max width=\textwidth}
\begin{tabular}{lccccc}
\toprule
\textbf{Feature / Capability}
& \textbf{PAN-PC-11~\cite{potthast2011pan}}
& \textbf{AraPlagDet 2015~\cite{bensalem2015araplagdet}}
& \textbf{PAWS~\cite{zhang2019paws}}
& \textbf{PAN 2026~\cite{pan2026generative}}
& \textbf{PlagiaCorpus (Ours)} \\
\midrule

Multilingual Support
& \checkmark
& $\times$ (Ar only)
& $\times$ (En only)
& $\times$ (En only)
& \checkmark \\
\midrule

Arabic Coverage
& $\times$
& \checkmark
& $\times$
& $\times$
& \checkmark \\
\midrule

Cross-Lingual Plagiarism
& \checkmark
& $\times$
& $\times$
& $\times$
& \checkmark \\
\midrule

Adversarial / Artificial Obfuscation
& \checkmark
& \checkmark
& \checkmark
& \checkmark
& \checkmark \\
\midrule

Complex Paraphrasing
& \checkmark
& \checkmark
& \checkmark
& \checkmark
& \checkmark \\
\midrule

LLM-Generated Plagiarism
& $\times$
& $\times$
& $\times$
& \checkmark
& \checkmark \\
\bottomrule
\end{tabular}
\end{adjustbox}
\end{table*}

\section*{Methods}

The PlagiaCorpus dataset was constructed through a multi-stage bilingual generation pipeline designed to represent progressively more challenging forms of text reuse in English and Arabic. The dataset combines two complementary components constructed from (i) Wikipedia content and (ii) book-length texts from two openly licensed collections, hereafter referred to as the \emph{book component}. These components were initially generated independently, subsequently merged, and finally curated into the released benchmark. As detailed in Table~\ref{tab:provenance}, the resulting corpus draws from English and Arabic Wikipedia, a public-domain English book collection, and a CC~BY~4.0-licensed Arabic book corpus, providing broad topical, linguistic, and stylistic coverage.

\begin{table}[!htbp]
\centering
\caption{Data provenance and licenses of the source repositories: passages initially collected and passages retained in the released source corpus after curation.}
\label{tab:provenance}
\begin{adjustbox}{max width=\textwidth}
\begin{tabular}{llrr}
\toprule
\textbf{Repository} & \textbf{License} & \textbf{Collected} & \textbf{Released (\%)} \\
\midrule
English Wikipedia & CC BY-SA 4.0 & 12,188 & 12,102 (34.69\%) \\
Arabic Wikipedia & CC BY-SA 4.0 & 6,549 & 6,483 (18.58\%) \\
The Arabic E-Book Corpus & CC BY 4.0 & 10,000 & 9,999 (28.66\%) \\
PleIAs English-PD & Public domain & 10,000 & 6,301 (18.06\%) \\
\midrule
\textbf{Total} & & \textbf{38,737} & \textbf{34,885 (100.00\%)} \\
\bottomrule
\end{tabular}
\end{adjustbox}
\end{table}

The construction methodology combines naturally occurring source material, deterministic text obfuscation, large language model (LLM)-based paraphrasing, cross-lingual translation with structural reformulation, and independently sampled non-reused texts. This design produces a controlled spectrum of text reuse ranging from exact lexical copying to cases in which the relationship between a query and its source is primarily semantic or cross-lingual.

\begin{figure*}[!htbp]
    \centering
    \includegraphics[width=\textwidth]{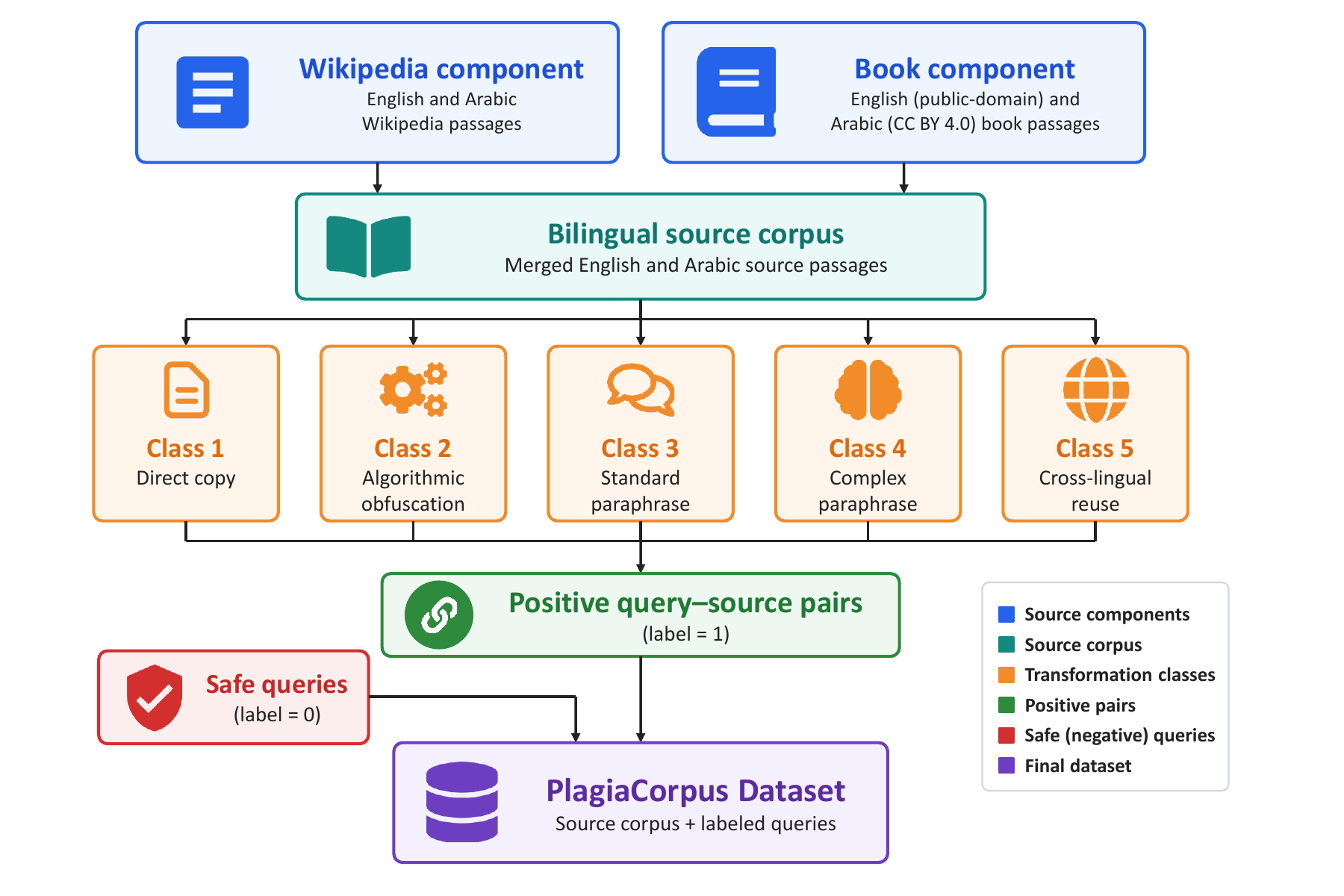}
    \caption{Overview of the PlagiaCorpus dataset construction pipeline. Source passages are collected from two complementary components, namely Wikipedia and openly licensed book collections, and merged into a unified bilingual source corpus. Five transformation strategies are then applied to generate increasingly challenging forms of text reuse, ranging from direct copying and algorithmic obfuscation to LLM-based paraphrasing and cross-lingual translation with structural reformulation. The resulting positive query--source pairs are combined with independently sampled safe queries, which serve as negative examples, to construct the final PlagiaCorpus benchmark.}
    \label{fig:PlagiaCorpus_pipeline}
\end{figure*}

Figure~\ref{fig:PlagiaCorpus_pipeline} summarizes the overall dataset construction process. English and Arabic source passages were first collected from the Wikipedia and book components and segmented into paragraph-level units. Each selected source passage was then transformed according to different reuse strategies to produce positive query-source pairs. Independently selected passages were additionally incorporated as negative Safe examples. Finally, the Wikipedia and book components, together with the generated transformations, were consolidated and curated into the final PlagiaCorpus dataset.

\subsection*{Source Corpus Construction}

The PlagiaCorpus source corpus was assembled from two complementary data components. The first component was constructed from English and Arabic Wikipedia content~\cite{wikimedia2024dumps}. A broad collection of topics spanning literature, history, culture, science, law, technology, philosophy, geography, and related domains was used to retrieve diverse Wikipedia articles in both languages through the MediaWiki API, using the \texttt{wikipedia} Python package, in April 2026. Article contents were segmented into paragraph-level passages using paragraph boundaries defined by the Wikipedia markup. To ensure that the retained passages contained sufficient contextual information for subsequent transformations, only paragraphs exceeding 200 characters were preserved, while short fragments and section headings were discarded.

The second component, the book component, was constructed from two book collections. English passages were obtained from the PleIAs English-PD collection~\cite{pleias2025englishpd}, which consists of public-domain texts. Arabic passages were obtained from The Arabic E-Book Corpus~\cite{hallberg2025arabicebookcorpus,hallberg2025arabicebookdata}, a collection of 1,745 books published by the Hindawi Foundation between 2008 and 2024 and distributed under a CC~BY~4.0 license. The Arabic corpus was accessed through a third-party mirror of the original deposit on the Hugging Face Hub (\texttt{mohres/The\_Arabic\_E-Book\_Corpus}). Paragraphs longer than 150 characters were streamed from each collection, and the first 10,000 paragraphs per language were retained as source passages. These resources complement Wikipedia with book-length material, thereby increasing the topical, stylistic, and structural diversity of the benchmark. Provenance and license information for each repository is provided in Table~\ref{tab:provenance}.

For both components, the selected English and Arabic passages formed the reference source corpus against which generated queries could be evaluated. The Wikipedia and book components were subsequently merged, producing a unified bilingual source collection covering heterogeneous topics, genres, and writing styles. Each selected source passage was subjected to transformations representing different forms and levels of text reuse. Table~\ref{tab:PlagiaCorpus_classes} summarizes these classes.

\begin{table*}[!htbp]
\centering
\caption{Text-reuse classes used in the PlagiaCorpus dataset.}
\label{tab:PlagiaCorpus_classes}
\begin{tabular}{p{1.4cm} p{2.5cm} p{3cm} p{7.2cm}}
\toprule
\textbf{Class} & \textbf{Reuse type} & \textbf{Generation method} & \textbf{Description} \\
\midrule

Class 1 &
Direct copy &
Identity transformation &
The source passage is reproduced verbatim, representing exact textual reuse and providing the least obfuscated positive examples. \\
\midrule
Class 2 &
Algorithmic modification &
Rule-based transformation &
The source is modified at the character level without changing its underlying content. English passages use leetspeak-style character substitutions, while Arabic passages are modified by inserting the Arabic tatweel (elongation) character (U+0640) within words. \\
\midrule
Class 3 &
Standard paraphrase &
Qwen2.5-14B-Instruct &
The source is rewritten in the same language using an instruction-tuned LLM while preserving its core meaning. This reduces direct lexical overlap while maintaining semantic correspondence with the source. \\
\midrule
Class 4 &
Complex paraphrase &
Gemma 4 31B &
The source is independently rewritten using a second LLM, introducing an alternative paraphrasing regime and additional variation in lexical choice and sentence structure. \\
\midrule
Class 5 &
Cross-lingual reuse &
Gemma 4 26B &
The source is translated between English and Arabic while simultaneously undergoing light structural paraphrasing. Both English-to-Arabic and Arabic-to-English directions are represented. \\
\midrule
Safe &
Non-reused text &
Independent sampling &
An independently selected passage that is not generated from the corresponding source material is used as a negative example, representing text for which no reuse relationship should be detected. \\

\bottomrule
\end{tabular}
\end{table*}

As shown in Table~\ref{tab:PlagiaCorpus_classes}, the five positive classes were designed to span a continuum of detection difficulty. Class~1 preserves complete lexical identity and therefore represents the most explicit form of reuse. Class~2 retains the original content but perturbs its surface representation, targeting methods that rely heavily on exact character or token overlap. Classes~3 and~4 weaken lexical and syntactic correspondence through LLM-based rewriting while maintaining the underlying semantics. Class~5 introduces an additional level of difficulty by transferring the source content across languages. The Safe category complements these positive classes with independently selected negative examples. Collectively, this structure enables PlagiaCorpus to evaluate lexical, semantic, and cross-lingual text-reuse detection within a common benchmark.

Table~\ref{tab:PlagiaCorpus_example} presents an illustrative example showing how the same source passage is transformed across the five positive classes while preserving a traceable relationship to the original text. The progression from exact copying (Class~1) to character-level obfuscation (Class~2), semantic paraphrasing using two different LLMs (Classes~3 and~4), and cross-lingual translation with structural reformulation (Class~5) illustrates the increasing difficulty posed to plagiarism and text-reuse detection systems. In contrast, the Safe example is sampled independently from the source corpus and shares no reuse relationship with the original passage, illustrating the negative examples included in the benchmark.

\begin{table*}[!htbp]
    \centering
    \includegraphics[width=\textwidth]{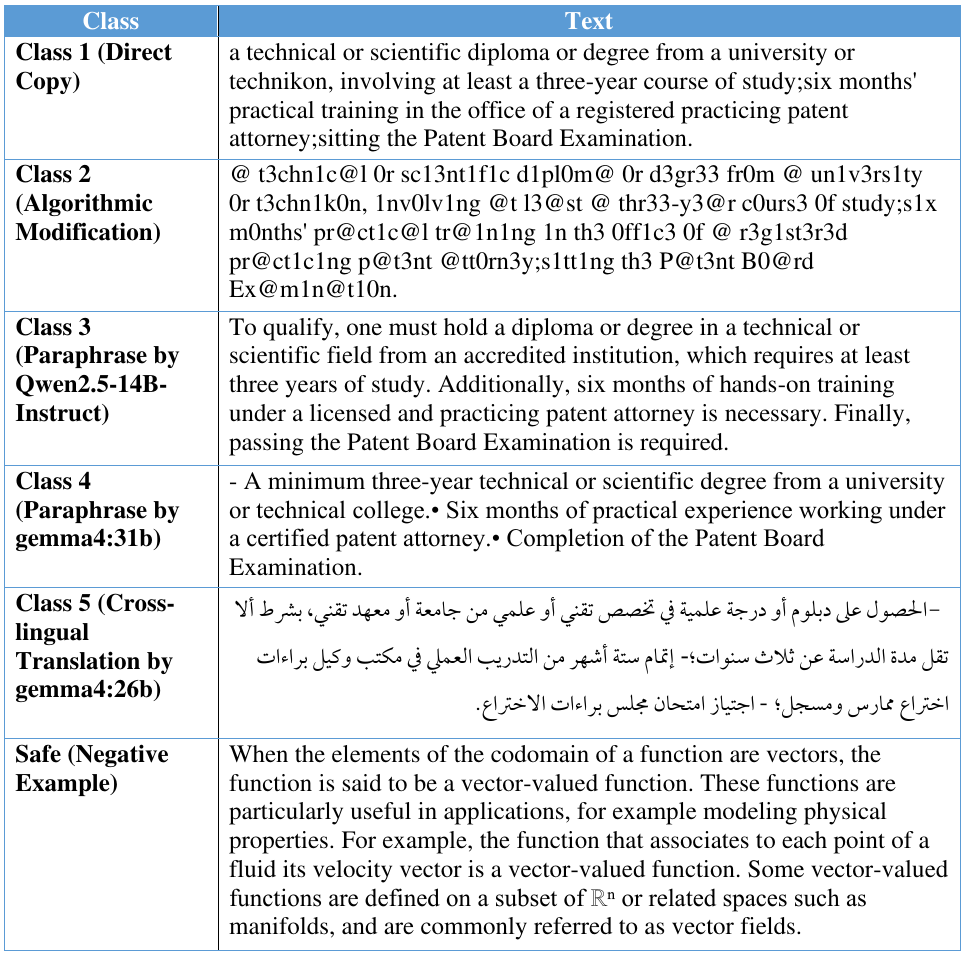}
    \caption{Illustrative example of a source passage (ID: \texttt{WIKI\_006573}) and its corresponding transformations across the five positive reuse classes in PlagiaCorpus, together with a Safe negative example. The examples demonstrate the progressive increase in transformation complexity from verbatim copying to algorithmic obfuscation, monolingual LLM-based rewriting, and cross-lingual semantic transfer.}
    \label{tab:PlagiaCorpus_example}
\end{table*}

\subsection*{Direct and Algorithmically Modified Reuse}

For \textbf{Class~1}, each selected source paragraph was copied without modification and used directly as a positive query. These samples represent explicit verbatim reuse and provide a baseline against which the more heavily transformed classes can be compared.

For \textbf{Class~2}, deterministic language-specific transformations were applied to alter the surface representation of the source while leaving its textual content recognizable. For English, lower-case Latin characters were replaced using leetspeak-style substitutions, specifically \texttt{e}$\rightarrow$\texttt{3}, \texttt{a}$\rightarrow$\texttt{@}, \texttt{i}$\rightarrow$\texttt{1}, and \texttt{o}$\rightarrow$\texttt{0}. For Arabic, the Arabic tatweel (elongation) character was inserted after the first character of every word containing more than two characters. These transformations create adversarial cases in which the passage remains readily interpretable to a human reader while its character-level representation differs from the original.

\subsection*{LLM-Based Paraphrase Generation}

Classes~3 and~4 represent semantic text reuse in which the meaning of the source passage is retained while its lexical and syntactic realization is changed. Two different LLMs were employed to diversify the generated paraphrases and reduce dependence on the rewriting characteristics of a single model.

For \textbf{Class~3}, \texttt{Qwen/Qwen2.5-14B-Instruct} was used to paraphrase source passages in their original language through Hugging Face Transformers in bf16 precision. Generation used stochastic decoding with a temperature of 0.7 and a limit of 256 new tokens. For the book component, top-$p$ was additionally set to 0.9. The model was explicitly instructed to rewrite the passage while preserving its core meaning and to return only the resulting paraphrase.

\begin{figure*}[!htbp]
    \centering
    \includegraphics[width=0.95\textwidth]{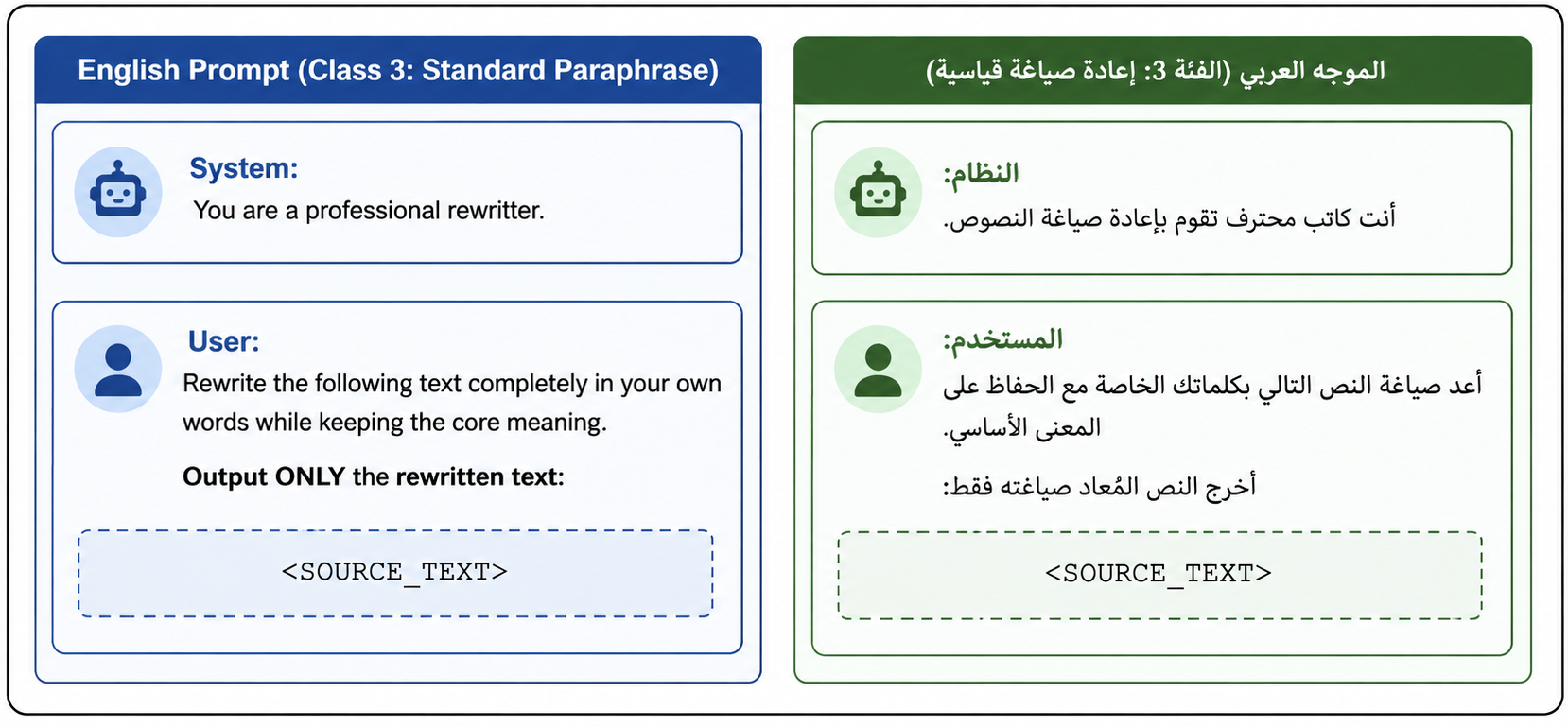}
    \caption{English and Arabic prompt templates used to generate Class~3 standard paraphrases with the \texttt{Qwen2.5-14B-Instruct} model. The prompts instruct the model to rewrite each source passage while preserving its core meaning and returning only the paraphrased text.}
    \label{fig:class3_prompt}
\end{figure*}

The English and Arabic prompt templates used for Class~3 paraphrase generation in the book component are illustrated in Figure~\ref{fig:class3_prompt}. Both prompts follow a system-user instruction format and direct the model to rewrite the source passage while preserving its core meaning and returning only the resulting paraphrased text. The same paraphrasing objective is maintained across both languages, with language-specific instructions provided for English and Arabic.

The Wikipedia component used the same paraphrasing objective with a more concise instruction. For English, the model was instructed to rewrite the passage completely in its own words and return only the rewritten text. The equivalent instruction was provided in Arabic for the Arabic source passages. Thus, although the exact prompt wording varied slightly between the two components of the dataset, Class~3 consistently represents same-language semantic paraphrasing using \texttt{Qwen2.5-14B-Instruct}.

For \textbf{Class~4}, an additional paraphrasing process was performed using \texttt{gemma4:31b} served through Ollama. The model was configured with a temperature of 0.7 and top-$p$ of 0.9 and instructed to rewrite the source while retaining its core meaning. Employing a second LLM provides an alternative paraphrase distribution, thereby increasing the diversity of semantic transformations represented in the benchmark. Moreover, the inclusion of paraphrases generated by two distinct LLMs allows plagiarism detection systems to be assessed for their robustness to model-specific paraphrasing styles and their ability to generalize across different generative models, rather than overfitting to the linguistic characteristics of a single LLM. Generation succeeded for 36,105 of the 38,737 collected passages; passages whose generation failed have no Class~4 query.

\begin{figure*}[!htbp]
    \centering
    \includegraphics[width=0.95\textwidth]{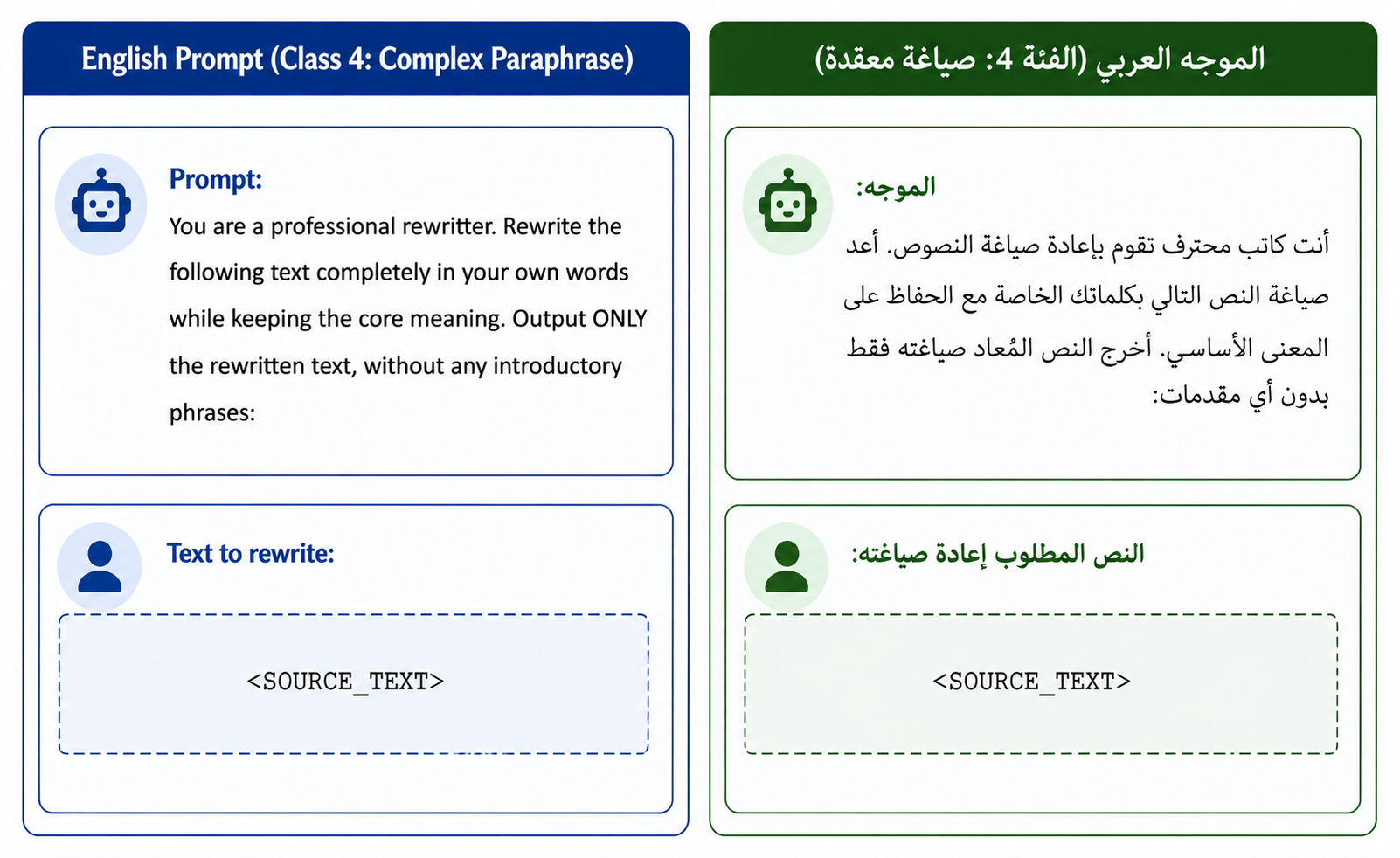}
    \caption{English and Arabic prompt templates used to generate Class~4 complex paraphrases with the \texttt{gemma4:31b} model. The prompts instruct the model to rewrite each source passage in its own words while preserving the core meaning and returning only the rewritten text without introductory content.}
    \label{fig:class4_prompt}
\end{figure*}

The English and Arabic prompt templates used for Class~4 paraphrase generation are illustrated in Figure~\ref{fig:class4_prompt}. In both languages, \texttt{gemma4:31b} was instructed to completely rewrite the source passage in its own words while preserving the underlying meaning. The prompts additionally constrain the model to return only the rewritten passage without introductory phrases, thereby reducing extraneous generated content and ensuring that the resulting samples consist solely of the transformed source text.

The use of two independently parameterized LLM-based paraphrasing strategies therefore provides complementary forms of semantic rewriting, allowing the benchmark to assess whether detection methods remain effective when lexical overlap with the original source is substantially reduced.

\subsection*{Cross-Lingual Text Reuse}

\textbf{Class~5} extends PlagiaCorpus from monolingual reuse to cross-lingual semantic reuse. Following construction and merging of the bilingual source corpus, the source passages were separated according to language and randomly sampled (seed 42) for a bidirectional English-Arabic transformation. Each source passage was used for at most one Class~5 query.

Cross-lingual samples were generated using \texttt{gemma4:26b} served through Ollama, for 6,870 English and 6,869 Arabic source passages. Importantly, the transformation was not formulated as literal translation alone. The model was explicitly instructed to produce natural target-language text while subtly reformulating its structure. Class~5 therefore represents cross-lingual text reuse in which the underlying semantic content is retained while both the language and surface realization are changed.

\begin{figure*}[!htbp]
    \centering
    \includegraphics[width=0.95\textwidth]{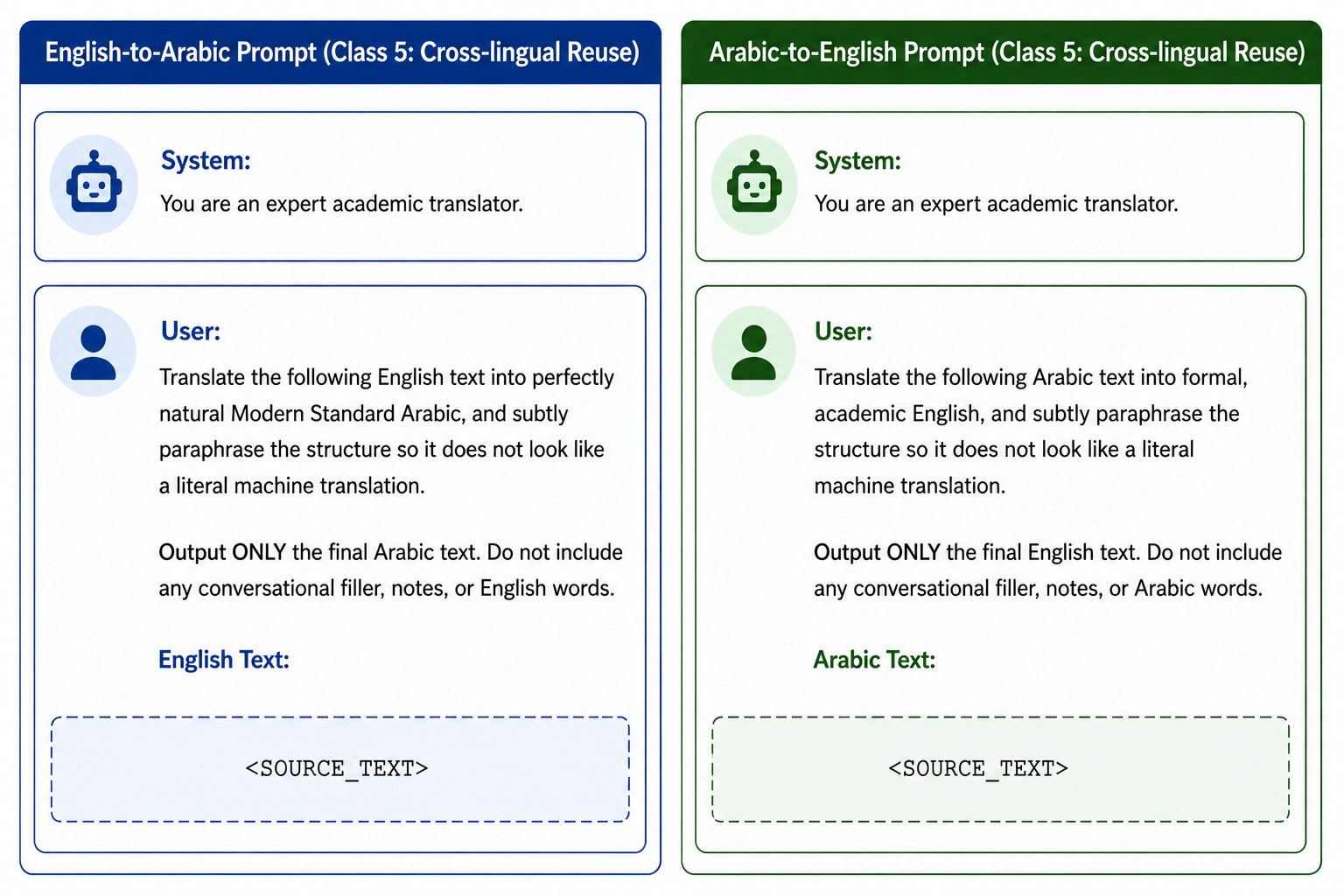}
    \caption{English-to-Arabic and Arabic-to-English prompt templates used to generate Class~5 cross-lingual text-reuse samples with the \texttt{gemma4:26b} model. The prompts combine translation with structural paraphrasing to preserve the source meaning while reducing literal correspondence between the source and target texts.}
    \label{fig:class5_prompt}
\end{figure*}

The bidirectional prompt templates used for Class~5 cross-lingual text-reuse generation are illustrated in Figure~\ref{fig:class5_prompt}. For English-to-Arabic transformation, \texttt{gemma4:26b} was instructed to produce natural Modern Standard Arabic while subtly paraphrasing the structure of the source passage. Conversely, Arabic-to-English transformation targeted formal academic English under the same structural paraphrasing constraint. In both directions, the prompts explicitly discourage literal translation and restrict the output to the target-language text only, thereby producing cross-lingual samples that preserve the underlying source meaning while reducing direct lexical and structural correspondence.

The two translation directions are labelled separately in the \texttt{type} field, while conceptually constituting a single cross-lingual reuse class. This class is particularly important for evaluating systems whose detection mechanisms extend beyond language-specific lexical overlap and instead rely on multilingual semantic correspondence.

\subsection*{Negative Control Samples}

A dedicated \textbf{Safe} category was incorporated to provide negative examples and prevent the benchmark from containing only reused or transformed material. Safe queries were not generated through any of the Class~1--5 transformations. Instead, they were independently selected from text pools separate from those used for the positive source passages.

For the Wikipedia component, separate English and Arabic control pools were crawled from articles retrieved for topics that were not used for the source passages. For the book component, Safe passages were taken from later portions of the English and Arabic book collections, after skipping the first 20,000 paragraphs of each collection, so that they do not overlap the source passages. Duplicate Safe passages were removed during curation (see below).

In the released dataset, 25\% of the Safe queries in each language (3,928 queries) were additionally transformed with the same character-level rules as Class~2. Without this step, the presence of leetspeak or tatweel would identify a query as positive. The released dataset contains 15,714 Safe queries assigned \texttt{label}=0. These samples provide a negative control because a detection system must distinguish genuine source-derived reuse from independently sourced text rather than simply assigning positive labels to every query.

\subsection*{Dataset Integration and Assembly}

The Wikipedia and book components were consolidated into a unified dataset after their respective source passages and transformation classes had been generated. Source passages received stable identifiers (\texttt{WIKI\_000000}, \ldots, \texttt{PD\_000000}, \ldots), and each positive query records the identifier of the passage from which it was derived. Positive examples from Classes~1--5 are assigned \texttt{label}=1, whereas Safe examples are assigned \texttt{label}=0. The original English and Arabic passages are retained as the reference source corpus, allowing each positive query to be evaluated against the material from which it was derived.

The final organization therefore combines heterogeneous source provenance with a controlled progression of reuse transformations. The Wikipedia component contributes broad encyclopedic and topic-driven coverage, while the book component introduces longer-form literary and non-fiction material from public-domain English books and contemporary Arabic books. Their integration produces a bilingual benchmark spanning exact copying, character-level obfuscation, monolingual semantic paraphrasing, cross-model paraphrasing, and cross-lingual semantic reuse, together with independently sourced negative examples.

\subsection*{Curation and Train/Test Split}

The assembled dataset was curated before release. All steps are deterministic (seed 42) and implemented in the code repository. The resulting release is available on Hugging Face~\cite{plagiacorpus2026data}. Table~\ref{tab:curation} summarizes the records removed at each step.

\textbf{Language filtering.} Arabic-script passages were accepted as Arabic. Latin-script passages were treated as English unless two signals agreed that they were not: fewer than 5\% English function words and a confidence below 0.2 for English according to the Lingua language identifier~\cite{lingua}. Source passages that were not English or Arabic, or whose language contradicted the language recorded in their queries, were removed together with all queries derived from them. Most removed passages were German (2,891) and French (792) text from the English book collection. Safe passages whose language differed from their label were also removed.

\textbf{LLM output cleaning.} For Classes~3--5 only, introductory phrases (e.g., ``Here is the rewritten text:''), trailing notes and markdown formatting were stripped. Outputs were removed when they were refusals, echoed the prompt, were shorter than 20 characters, were in the wrong language or script, contained characters from an unrelated script (most often Chinese characters produced by Qwen2.5 when paraphrasing Arabic), or were identical to their source. Outputs that end without terminal punctuation while their source does (9,268 queries), or that contain fewer than 30\% of the source's words (1,025 queries), are retained but flagged in the \texttt{quality\_flags} field.

\textbf{De-duplication.} Safe queries identical to a source passage or to another Safe query were removed. In addition, all source and Safe passages were embedded with LaBSE~\cite{feng2022labse}, and pairs with a cosine similarity of at least 0.85 were identified. Safe passages matching a source passage at this threshold, including translations of a source passage, were removed. Source passages with an exact duplicate (3,393 passages in 1,393 groups) or a near-duplicate (5,819 pairs, 436 of them cross-lingual) were retained but treated as a single group for splitting.

\textbf{Grouped split.} Each source passage, merged with its duplicates and near-duplicates, defines a group, and every query derived from a group is assigned to the same split. Groups were assigned to the test split within strata defined by collection and language (Safe queries form their own strata) until 10\% of the queries of each stratum were reached. As a result, no source passage contributes queries to both the training and the test split.

\begin{table}[!htbp]
\centering
\caption{Records removed during curation.}
\label{tab:curation}
\begin{adjustbox}{max width=\textwidth}
\begin{tabular}{lr}
\toprule
\textbf{Reason} & \textbf{Records} \\
\midrule
\multicolumn{2}{l}{\textit{Source corpus}} \\
Passage not in English or Arabic, or language inconsistent with its queries & 3,852 \\
\midrule
\multicolumn{2}{l}{\textit{Queries}} \\
Derived from a removed source passage & 15,951 \\
Safe query in a language other than its label & 5,920 \\
Safe query duplicating another Safe query & 14,705 \\
Safe query identical to a source passage & 2,330 \\
Safe query near-duplicate or translation of a source passage (LaBSE) & 68 \\
LLM output containing characters of an unrelated script & 1,756 \\
LLM output in the wrong language or script & 284 \\
LLM refusal & 20 \\
\midrule
\textbf{Total queries removed} & \textbf{41,034} \\
\bottomrule
\end{tabular}
\end{adjustbox}
\end{table}

\section*{Data Records}

The PlagiaCorpus dataset is hosted on Hugging Face~\cite{plagiacorpus2026data} and is organized into two complementary components: a \textit{source corpus} containing the reference passages and a \textit{queries} collection containing transformed and negative examples used for benchmarking plagiarism and text-reuse detection systems. In total, the dataset comprises 198,643 text records. The source corpus contains 34,885 reference passages totaling 5.27 million words, while the remaining 163,758 records are queries, comprising 147,381 training queries and 16,377 test queries. The corpus integrates content from Wikipedia and two book collections, as summarized in Table~\ref{tab:provenance}.

The dataset consists of three files, each provided in JSON and in Parquet format: \path{PlagiaCorpus_corpus} (source passages), \path{PlagiaCorpus_queries_train} and \path{PlagiaCorpus_queries_test}. After language filtering, every record is in English or Arabic (Table~\ref{tab:languages}).

\begin{table}[!htbp]
\centering
\caption{Language distribution across all text records (source passages and queries).}
\label{tab:languages}
\begin{tabular}{lrr}
\toprule
\textbf{Language} & \textbf{Count} & \textbf{Percentage} \\
\midrule
English & 103,269 & 51.99\% \\
Arabic & 95,374 & 48.01\% \\
\midrule
\textbf{Total} & \textbf{198,643} & \textbf{100.00\%} \\
\bottomrule
\end{tabular}
\end{table}

The source corpus and query collection adopt complementary record structures, summarized in Table~\ref{tab:data_records}. Each source corpus record contains a unique identifier, the reference passage, its language, and provenance and license metadata. Query records additionally describe the reuse class, the language of the query and of its source, the transformation applied, the binary label, the generating model, quality flags, and the identifier of the source passage.

\begin{table*}[!htbp]
\centering
\footnotesize
\renewcommand{\arraystretch}{1.15}
\caption{Fields contained in the PlagiaCorpus source corpus and query records (C: corpus; Q: queries).}
\label{tab:data_records}
\begin{tabularx}{\textwidth}{@{}>{\ttfamily}l c >{\raggedright\arraybackslash}X@{}}
\toprule
\textrm{\textbf{Field}} & \textbf{In} & \textbf{Description} \\
\midrule
\_id & C, Q & Unique identifier: \texttt{WIKI\_000000} or \texttt{PD\_000000} for source passages, \texttt{Q\_000000} for queries. \\
text & C, Q & Text of the source passage or query. \\
language & C, Q & Language of the text (\texttt{en} or \texttt{ar}). \\
domain & C, Q & Source collection, e.g., \textit{Wikipedia (Wikimedia Foundation), accessed April 2026}, \textit{The Arabic E-Book Corpus (Hindawi Foundation)}, \textit{PleIAs English-PD}. \\
source & C, Q & Attribution: Wikipedia article URL or book-collection citation with book title; for queries, preceded by the transformation applied. \\
source\_license & C, Q & License of the underlying text (CC BY-SA 4.0, CC BY 4.0, or public domain). \\
\midrule
type & Q & Reuse class and language, e.g., \texttt{Class 4 (AR)}, \texttt{Class 5 (EN->AR)}. \\
reuse\_class & Q & \texttt{Class 1}--\texttt{Class 5}, or \texttt{Safe}. \\
source\_language & Q & Language of the source passage (\texttt{null} for Safe queries). \\
transformation & Q & \texttt{verbatim}, \texttt{algorithmic}, \texttt{llm\_paraphrase}, \texttt{llm\_cross\_lingual}, or \texttt{none}; obfuscated Safe queries are marked \texttt{algorithmic}. \\
label & Q & 1 for reused content (Classes 1--5), 0 for Safe queries. \\
generator\_model & Q & Generating model (Qwen2.5-14B-Instruct, gemma4:31b, gemma4:26b), or rule-based/human origin. \\
original\_text\_id & Q & Identifier of the source passage of the query (\texttt{null} for Safe queries). \\
quality\_flags & Q & Automatic flags for LLM outputs: \texttt{possibly\_truncated}, \texttt{much\_shorter\_than\_source}. \\
\bottomrule
\end{tabularx}
\end{table*}

The \texttt{original\_text\_id} field establishes an explicit correspondence between each positive query and its source passage, enabling reproducible evaluation of retrieval-based plagiarism detection systems. Additional metadata, including the reuse class, transformation, languages, binary label and generation model, supports fine-grained analysis of detector performance across text-reuse strategies, language pairs, source collections and LLMs.

\subsection*{License}
PlagiaCorpus is released under the CC~BY-SA~4.0 license (Creative Commons Attribution-ShareAlike 4.0 International). This license is required because the source corpus and the derived queries include Wikipedia text, which is distributed under CC~BY-SA~4.0. Passages from The Arabic E-Book Corpus are redistributed under the attribution terms of its CC~BY~4.0 license, which permits inclusion in CC~BY-SA~4.0 adaptations, and passages from PleIAs English-PD are in the public domain. Attribution is provided at the record level through the \texttt{domain}, \texttt{source} and \texttt{source\_license} fields. Redistributions of PlagiaCorpus and of adapted versions must retain this attribution and be released under the same license.

\section*{Technical Validation}
The released query set comprises 148,044 positive and 15,714 Safe queries (90.4\% and 9.6\%). Table~\ref{tab:topology} reports the number of queries and word-length statistics for each class, with words counted as whitespace-separated tokens. Table~\ref{tab:split} gives the number of queries per class in each split. The grouped split was verified automatically: no source passage, and no identical query text, occurs in both splits.

\begin{table*}[!htbp]
\centering
\caption{Number of queries and word-length statistics (whitespace tokens) for each text-reuse class.}
\label{tab:topology}
\begin{adjustbox}{max width=\textwidth}
\begin{tabular}{lrlrrr}
\toprule
\textbf{Class} & \textbf{Count (\%)} & \textbf{Generator} & \textbf{Mean} & \textbf{Median} & \textbf{Std Dev} \\
\midrule
Class 1 (EN) & 18,403 (11.2\%) & None (direct copy) & 167.7 & 120 & 151.2 \\
Class 1 (AR) & 16,482 (10.1\%) & None (direct copy) & 132.7 & 88 & 137.3 \\
Class 2 (EN) & 18,403 (11.2\%) & None (rule-based) & 167.7 & 120 & 151.2 \\
Class 2 (AR) & 16,482 (10.1\%) & None (rule-based) & 132.7 & 88 & 137.3 \\
Class 3 (EN) & 18,252 (11.1\%) & Qwen2.5-14B-Instruct & 112.8 & 106 & 58.9 \\
Class 3 (AR) & 14,679 (9.0\%) & Qwen2.5-14B-Instruct & 89.3 & 96 & 32.2 \\
Class 4 (EN) & 17,426 (10.6\%) & Gemma 4 31B & 122.3 & 92 & 94.2 \\
Class 4 (AR) & 15,068 (9.2\%) & Gemma 4 31B & 104.1 & 79 & 77.7 \\
Class 5 (EN$\rightarrow$AR) & 5,997 (3.7\%) & Gemma 4 26B & 162.4 & 121 & 137.9 \\
Class 5 (AR$\rightarrow$EN) & 6,852 (4.2\%) & Gemma 4 26B & 195.6 & 142 & 163.8 \\
Safe (EN) & 5,530 (3.4\%) & None (human text) & 104.9 & 70 & 101.9 \\
Safe (AR) & 10,184 (6.2\%) & None (human text) & 75.9 & 56 & 59.6 \\
\bottomrule
\end{tabular}
\end{adjustbox}
\end{table*}

\begin{table}[!htbp]
\centering
\caption{Number of queries per class in the training and test splits.}
\label{tab:split}
\begin{tabular}{lrr}
\toprule
\textbf{Class} & \textbf{Train} & \textbf{Test} \\
\midrule
Class 1 (EN) & 16,593 & 1,810 \\
Class 1 (AR) & 14,807 & 1,675 \\
Class 2 (EN) & 16,593 & 1,810 \\
Class 2 (AR) & 14,807 & 1,675 \\
Class 3 (EN) & 16,454 & 1,798 \\
Class 3 (AR) & 13,164 & 1,515 \\
Class 4 (EN) & 15,704 & 1,722 \\
Class 4 (AR) & 13,566 & 1,502 \\
Class 5 (EN$\rightarrow$AR) & 5,407 & 590 \\
Class 5 (AR$\rightarrow$EN) & 6,144 & 708 \\
Safe (EN) & 4,977 & 553 \\
Safe (AR) & 9,165 & 1,019 \\
\midrule
\textbf{Total} & \textbf{147,381} & \textbf{16,377} \\
\bottomrule
\end{tabular}
\end{table}

Paraphrases are shorter than their sources: on average, Class~3 queries contain 33\% (English) and 33\% (Arabic) fewer words than Class~1 queries, and Class~4 queries 27\% and 22\% fewer. LLM outputs that end without terminal punctuation, which may occur for long passages given the limit of 256 new tokens for Class~3, are identified in the \texttt{quality\_flags} field.

\subsection*{Human Annotation Reliability}

To assess the quality of the generated queries, a sample of 300 queries (50 per class, Classes~1--5 and Safe) was drawn from the dataset before curation. Each query was shown together with its source passage (for Safe queries, an unrelated passage from the same component and language) and rated independently by two annotators, who did not see the class label, on three ordinal criteria: \textit{Linguistic Quality} (1--5), \textit{Semantic Preservation} (1--5), and \textit{Obfuscation Difficulty} (1--3). Inter-annotator agreement was evaluated using quadratic weighted Cohen's $\kappa$, together with the percentage of ratings within one scale point (Within $\pm1$).

Table~\ref{tab:iia} summarizes the agreement statistics. Semantic Preservation achieved almost perfect agreement ($\kappa=0.921$), while Linguistic Quality ($\kappa=0.730$) and Obfuscation Difficulty ($\kappa=0.739$) exhibited substantial agreement. The Within $\pm1$ agreement exceeded 88\% for all dimensions, reaching over 95\% for Obfuscation Difficulty. These results indicate that the annotation protocol yields consistent quality assessments.

\begin{table}[h]
\centering
\caption{Inter-annotator agreement on the manually annotated validation sample (300 queries). For each dimension, items that either annotator left blank or marked as not applicable are excluded.}
\label{tab:iia}
\begin{tabular}{lcc}
\hline
\textbf{Annotation Dimension} & \textbf{Weighted $\kappa$} & \textbf{Within $\pm1$} \\
\hline
Linguistic Quality        & 0.730 & 88.67\% \\
Semantic Preservation     & 0.921 & 94.31\% \\
Obfuscation Difficulty    & 0.739 & 95.74\% \\
\hline
\end{tabular}
\end{table}

\section*{Usage Notes}
PlagiaCorpus is distributed in JSON and Parquet formats, which integrate directly with retrieval-augmented generation (RAG) and dense embedding workflows. Researchers should use \path{PlagiaCorpus_corpus} as the static index for vector databases (e.g., FAISS, ChromaDB). \path{PlagiaCorpus_queries_train} (90\%) and \path{PlagiaCorpus_queries_test} (10\%) are designed for training and evaluating source retrieval (e.g., MRR@10, Recall@$K$ over positive queries) and query--source pair verification (e.g., with cross-encoders). Because the split is grouped by source passage, test queries cannot be matched by memorizing training queries derived from the same passage. Users may exclude queries carrying \texttt{quality\_flags} for stricter evaluation.

The contents of PlagiaCorpus were collected through automated extraction from Wikipedia and openly licensed book collections and may therefore contain diverse political, ethical, cultural, religious, or other viewpoints expressed in the underlying source material. Such content is included solely as part of the source material and does not represent the views or positions of the authors, their institutions, or the funding organizations. PlagiaCorpus is intended for research on plagiarism and text-reuse detection; its use is governed by the CC~BY-SA~4.0 license described in Data Records.

\section*{Code Availability}
All Python code used for dataset generation, algorithmic obfuscation, LLM prompting, curation, splitting and statistical validation is available in the GitHub repository \url{https://github.com/riotu-lab/PlagiaCorpus}.

\section*{Author Contributions}
A.A. conceived and designed the study, generated the dataset and developed the accompanying code, performed the technical validation and statistical analysis, and drafted the manuscript.\\
S.S. contributed to the investigation, generated the dataset, and performed the technical validation and statistical analysis.\\
W.B. contributed to the technical validation, supervised the work, and critically revised the manuscript.\\
All authors approved the submitted version and agree to be personally accountable for their own contributions and to ensure that questions related to the accuracy or integrity of any part of the work are appropriately investigated and resolved.

\section*{Competing Interests}
The authors declare that they have no competing financial or non-financial interests that are relevant to the work described.

\section*{Acknowledgments}
This research was conducted with the support of the Cultural Research Grants Program launched by the Ministry of Culture in the Kingdom of Saudi Arabia. All views expressed in this paper are those of the authors and do not necessarily reflect the views of the Ministry. The authors also acknowledge the support of Prince Sultan University for paying the Article Processing Charges (APC) of this publication.

\end{document}